\documentclass[journal]{IEEEtran}
\usepackage{cite}
\ifCLASSINFOpdf
  \usepackage[pdftex]{graphicx}
\else
\fi
\usepackage{amsmath}
\usepackage{bbm}
\usepackage{booktabs}

\ifCLASSOPTIONcompsoc
 \usepackage[caption=false,font=normalsize,labelfont=sf,textfont=sf]{subfig}
\else
 \usepackage[font=footnotesize]{subfig}
\fi
\usepackage{url}
\begin{document}
%
% paper title
% Titles are generally capitalized except for words such as a, an, and, as,
% at, but, by, for, in, nor, of, on, or, the, to and up, which are usually
% not capitalized unless they are the first or last word of the title.
% Linebreaks \\ can be used within to get better formatting as desired.
% Do not put math or special symbols in the title.
\title{Semi-weakly Supervised Contrastive Representation Learning for Retinal Fundus Images}
%
%
% author names and IEEE memberships
% note positions of commas and nonbreaking spaces ( ~ ) LaTeX will not break
% a structure at a ~ so this keeps an author's name from being broken across
% two lines.
% use \thanks{} to gain access to the first footnote area
% a separate \thanks must be used for each paragraph as LaTeX2e's \thanks
% was not built to handle multiple paragraphs
%

\author{Boon~Peng~Yap,
        Beng~Koon~Ng% <-this % stops a space
\thanks{The computational work for this article was fully performed on resources of the National Supercomputing Centre, Singapore (https://www.nscc.sg).}% <-this % stops a space
\thanks{BP. Yap and BK. Ng are with the School of Electrical and Electronic Engineering, Nanyang Technological University, Singapore.}% <-this % stops a space
% \thanks{Manuscript received April 19, 2005; revised August 26, 2015.}
}

% note the % following the last \IEEEmembership and also \thanks - 
% these prevent an unwanted space from occurring between the last author name
% and the end of the author line. i.e., if you had this:
% 
% \author{....lastname \thanks{...} \thanks{...} }
%                     ^------------^------------^----Do not want these spaces!
%
% a space would be appended to the last name and could cause every name on that
% line to be shifted left slightly. This is one of those "LaTeX things". For
% instance, "\textbf{A} \textbf{B}" will typeset as "A B" not "AB". To get
% "AB" then you have to do: "\textbf{A}\textbf{B}"
% \thanks is no different in this regard, so shield the last } of each \thanks
% that ends a line with a % and do not let a space in before the next \thanks.
% Spaces after \IEEEmembership other than the last one are OK (and needed) as
% you are supposed to have spaces between the names. For what it is worth,
% this is a minor point as most people would not even notice if the said evil
% space somehow managed to creep in.

% The paper headers
% \markboth{Journal of \LaTeX\ Class Files,~Vol.~14, No.~8, August~2015}%
\markboth{Preprint}%
{Shell \MakeLowercase{\textit{et al.}}: Bare Demo of IEEEtran.cls for IEEE Journals}
% The only time the second header will appear is for the odd numbered pages
% after the title page when using the twoside option.
% 
% *** Note that you probably will NOT want to include the author's ***
% *** name in the headers of peer review papers.                   ***
% You can use \ifCLASSOPTIONpeerreview for conditional compilation here if
% you desire.

% If you want to put a publisher's ID mark on the page you can do it like
% this:
%\IEEEpubid{0000--0000/00\$00.00~\copyright~2015 IEEE}
% Remember, if you use this you must call \IEEEpubidadjcol in the second
% column for its text to clear the IEEEpubid mark.

% use for special paper notices
%\IEEEspecialpapernotice{(Invited Paper)}

% make the title area
\maketitle

% As a general rule, do not put math, special symbols or citations
% in the abstract or keywords.
\begin{abstract}
We explore the value of weak labels in learning transferable representations for medical images. Compared to hand-labeled datasets, weak or inexact labels can be acquired in large quantities at significantly lower cost and can provide useful training signals for data-hungry models such as deep neural networks. We consider weak labels in the form of pseudo-labels and propose a semi-weakly supervised contrastive learning (SWCL) framework for representation learning using semi-weakly annotated images. Specifically, we train a semi-supervised model to propagate labels from a small dataset consisting of diverse image-level annotations to a large unlabeled dataset. Using the propagated labels, we generate a patch-level dataset for pretraining and formulate a multi-label contrastive learning objective to capture position-specific features encoded in each patch. We empirically validate the transfer learning performance of SWCL on seven public retinal fundus datasets, covering three disease classification tasks and two anatomical structure segmentation tasks. Our experiment results suggest that, under very low data regime, large-scale ImageNet pretraining on improved architecture remains a very strong baseline, and recently proposed self-supervised methods falter in segmentation tasks, possibly due to the strong invariant constraint imposed. Our method surpasses all prior self-supervised methods and standard cross-entropy training, while closing the gaps with ImageNet pretraining.
\end{abstract}

% Note that keywords are not normally used for peerreview papers.

% For peer review papers, you can put extra information on the cover
% page as needed:
% \ifCLASSOPTIONpeerreview
% \begin{center} \bfseries EDICS Category: 3-BBND \end{center}
% \fi
%
% For peerreview papers, this IEEEtran command inserts a page break and
% creates the second title. It will be ignored for other modes.
\IEEEpeerreviewmaketitle

\section{Introduction}
% The very first letter is a 2 line initial drop letter followed
% by the rest of the first word in caps.
% 
% form to use if the first word consists of a single letter:
% \IEEEPARstart{A}{demo} file is ....
% 
% form to use if you need the single drop letter followed by
% normal text (unknown if ever used by the IEEE):
% \IEEEPARstart{A}{}demo file is ....
% 
% Some journals put the first two words in caps:
% \IEEEPARstart{T}{his demo} file is ....
% 
% Here we have the typical use of a "T" for an initial drop letter
% and "HIS" in caps to complete the first word.
% \IEEEPARstart{T}{his} demo file is intended to serve as a ``starter file''
% for IEEE journal papers produced under \LaTeX\ using
% IEEEtran.cls version 1.8b and later.
% % You must have at least 2 lines in the paragraph with the drop letter
% % (should never be an issue)
% I wish you the best of success.

% \hfill mds
 
% \hfill August 26, 2015

\IEEEPARstart{T}{ransfer} learning via the pretraining and fine-tuning paradigm is one of the most popular approaches in training deep convolutional neural networks (CNN) for medical imaging tasks, e.g., diabetic retinopathy detection \cite{gulshan2016development}. In a typical transfer learning setting, a CNN is first initialized with weights pretrained on a large generic dataset, after which the weights are fine-tuned to solve a specific task on a smaller dataset. Although there might be domain mismatch between the pretrain dataset and the fine-tune dataset, the pretrained weights can still provide a good starting point for fine-tuning; in most cases models initialized with pretrained weights are able to converge faster and require less annotated samples compared to their randomly initialized counterparts.

Traditionally, pretraining is performed in a fully supervised manner on a large hand-labeled dataset such as ImageNet \cite{Deng2009ImageNetAL}. Recently there is a push for self-supervised pretraining methods which seeks to extract generalizable representations without requiring human-annotated samples. Notable examples from the natural image domain include relative position prediction \cite{Doersch2015UnsupervisedVR}, image inpainting \cite{pathak2016context}, rotation angle prediction \cite{gidaris2018unsupervised}, contrastive learning \cite{chen2020simple, He2020MomentumCF}, and other consistency-based methods \cite{grill2020bootstrap, Zbontar2021BarlowTS, Caron2021EmergingPI}. SimCLR \cite{chen2020simple} popularizes the contrastive learning objective for self-supervised pretraining on image data. On several classification datasets, models fine-tuned on representations learned by SimCLR outperform or match the performance of supervised ImageNet pretraining. In the medical imaging domain, Model Genesis \cite{zhou2021models} introduces a reconstruction-based objective for 3D medical images and achieves superior performance on various 3D image segmentation tasks compared to 2D models pretrained on ImageNet.

% An example of a floating figure using the graphicx package.
% Note that \label must occur AFTER (or within) \caption.
% For figures, \caption should occur after the \includegraphics.
% Note that IEEEtran v1.7 and later has special internal code that
% is designed to preserve the operation of \label within \caption
% even when the captionsoff option is in effect. However, because
% of issues like this, it may be the safest practice to put all your
% \label just after \caption rather than within \caption{}.
%
% Reminder: the "draftcls" or "draftclsnofoot", not "draft", class
% option should be used if it is desired that the figures are to be
% displayed while in draft mode.
%
\begin{figure}[!t]
\centering
\includegraphics[width=3.45in]{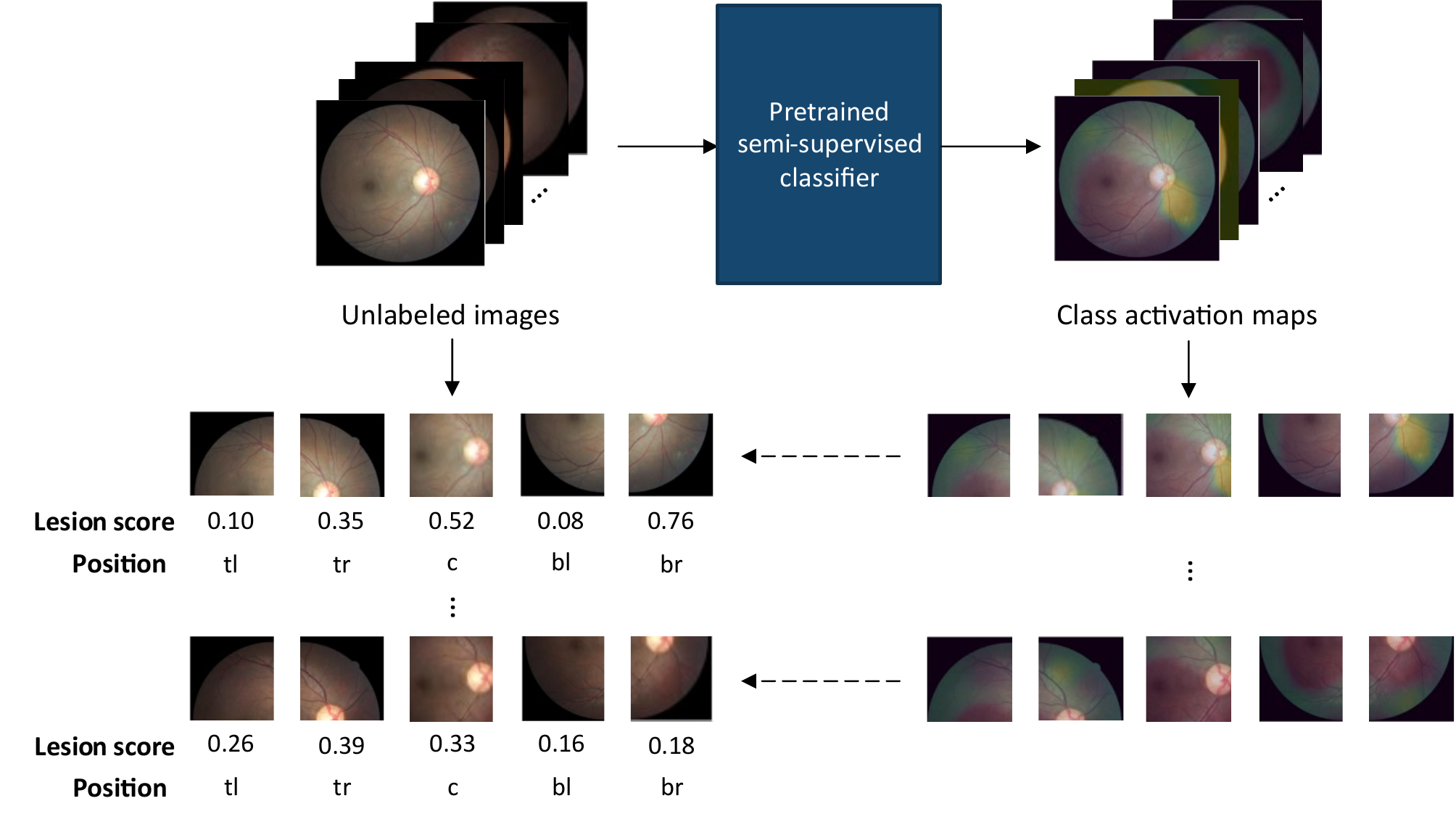}
\caption{Generating a semi-weakly annotated patch dataset from unlabeled images, using CAMs extracted from a semi-supervised abnormality classifier. Each image and its corresponding CAM are cropped into five fixed patches: top-left (tl), top-right (tr), center (c), bottom-left (bl), and bottom-right (br). The lesion score of each image patch is computed as the average activation value of the normalized and cropped CAM.}
\label{fig_gen}
\end{figure}

% Note that the IEEE typically puts floats only at the top, even when this
% results in a large percentage of a column being occupied by floats.

In between fully supervised and self-supervised learning, there is another training paradigm known as weakly-supervised learning, which involve training on weak labels obtained from potentially noisy source. Examples of weak labels include labels extracted from clinical reports accompanied by medical images \cite{Wang2017ChestXRay8HC, zhang2020contrastive}, and pseudo-labels generated by models pretrained on small hand-labeled datasets. In contrast to expert annotations which are prohibitively expensive to acquire, weak labels can be obtained in large quantities with minimal manual effort. In this work, we explore the efficacy of weakly-annotated labels as training signals for generic representation learning. In particular, we propose a \textbf{S}emi-\textbf{W}eakly supervised \textbf{C}ontrastive \textbf{L}earning (SWCL) framework for learning transferable visual representations from a semi-weakly annotated fundus patch dataset. We first propagate labels from OIA-ODIR \cite{Li2020ABO}, a relatively small dataset consisting of image-level annotations for a wide variety of retinal diseases, to a large set of fundus images from the Kaggle-EyePACS \cite{Cuadros2009EyePACSAA} dataset through semi-supervised learning. Kaggle-EyePACS is one of the largest public datasets for diabetic retinopathy (DR) grading. Despite being labeled for DR, the images in this dataset contain other types of diseases that have not been explicitly labeled by the dataset provider \cite{Wang2019RetinalAR}. Therefore, we treat the entire Kaggle-EyePACS dataset as unlabeled and use it along with the labeled OIA-ODIR dataset to train a semi-supervised pseudo-labeler. After the pseudo-labeler is trained, we extract the class activation maps (CAMs) \cite{zhou2016learning} for every image in both labeled and unlabeled datasets to obtain a set of image-CAM pairs which will be used to construct a patch-level dataset for pretraining.

To learn spatially consistent representations with position-specific lesion information, we construct a \textbf{S}emi-\textbf{W}eakly \textbf{A}nnotated \textbf{P}atch dataset for retinal images (retinal-SWAP) from the image-CAM pairs. As shown in Figure \ref{fig_gen}, we crop each image and its CAM into five fixed patches and assign each patch a lesion score and a (relative) position label. The patch-wise lesion scores are computed from the crops of CAMs while the position labels are the relative positions of the crops with respect to the whole image. For normal (healthy) images from OIA-ODIR, we infer the lesion scores directly from the ground-truth labels, i.e., setting them to zero, while the rest of the patches depends on the activation values of the CAMs extracted from the pseudo-labeler, hence the constructed dataset is \textit{semi-weakly} annotated.

Using retinal-SWAP as pretrain dataset, we propose a new training objective based on the supervised contrastive learning objective \cite{NEURIPS2020_d89a66c7}. The training objective encourages the model to learn representations such that patches with the same labels are pulled together in a projected vector space, while patches with mismatched labels are pushed away from each other. Each patch in the pretrain dataset contains retinal structures specific to each relative position. For example, optic discs and macula are typically concentrated at the center, while blood vessel endings are found towards the edge of the fundus images. We hypothesize that learning to identify lesion regions from similar background anatomical structures via contrastive learning would benefit both downstream classification and segmentation tasks. We empirically verify this hypothesis on seven public retinal fundus datasets and compare the transfer learning performance with self-supervised pretraining, standard in-domain cross-entropy pretraining, and large-scale supervised pretraining methods. We release the codes and pretrained models at \url{https://github.com/BPYap/SWCL}.

Our contributions in this paper are three-fold:
\begin{enumerate}
    \item We demonstrate how to construct a semi-weakly annotated patch dataset for pretraining by propagating image-level annotations from a small dataset to a large unlabeled dataset in the form of class activation maps.
    \item We formulate a multi-label contrastive learning objective to enforce consistency between patches with similar lesion and position information. 
    \item 	We conduct extensive transfer learning and ablation experiments to assess the effectiveness of our approach. We perform benchmarking on seven retinal fundus datasets that cover a wide range of tasks, including joint coarse- and fine-grained diabetic retinopathy classification, glaucoma classification, retinal vessel segmentation, and joint optic disc and optic cup segmentation. Our approach outperforms state-of-the-art self-supervised methods and standard cross-entropy objective while being competitive with large-scale ImageNet pretraining.
\end{enumerate}

\section{Related Work}
In this section, we review previous works on two common paradigms of pretraining deep neural networks: supervised and self-supervised pretraining.

\subsection{Supervised pretraining}
Supervised pertaining involves training a model on large generic hand-labeled datasets, with anticipations that the model learns useful representations that can generalize to other tasks. Models pretrained on the ImageNet dataset \cite{Deng2009ImageNetAL}, usually on the smaller ILSVRC-2012 split \cite{Russakovsky2015ImageNetLS}, has been extensively utilized in the medical imaging literature. Some applications include detection of COVID-19 \cite{Wang2020COVIDNetAT} and bone age estimation \cite{Lee2017FullyAD} on X-rays, lung disease classification \cite{Shin2016DeepCN} and organ segmentation \cite{ Man2019DeepQL} on CT scans, breast cancer detection \cite{Carneiro2017AutomatedAO, Wu2020DeepNN} on mammograms, melanoma classification \cite{Brinker2019ACN, Liu2020ADL} on dermoscopic images, and DR \cite{gulshan2016development} and glaucoma \cite{Orlando2020REFUGECA} classification on retinal fundus images. Prior empirical observations show that weights pretrained on ImageNet provide a good starting point for solving downstream tasks; they greatly speed up model convergence and can even match or surpass human-level performance after fine-tuning on small hand-labeled datasets.

Following the recent advances in network architecture design and training algorithms, Kolesnikov, Beyer \textit{et al.} \cite{Kolesnikov2020BigT} revisit supervised pretraining on ImageNet and propose BiT, a series of models pretrained on a modified version of the ResNet \cite{he2016deep} architecture. Our work in this paper is based on this modified ResNet architecture and we benchmark against the public release of the pretrained BiT models in the experiment section. On the other hand, Khosla \textit{et al.} \cite{NEURIPS2020_d89a66c7} extends self-supervised contrastive learning objective (discussed in the next section) to the supervised setting. The supervised contrastive learning objective were experimentally shown to outperform the standard cross-entropy objective in terms of classification accuracy and robustness to noise.

\subsection{Self-supervised pretraining}
Research in self-supervised learning has gained a lot of popularity in recent years. This approach aims to learn representations from a vast number of unlabeled samples using labels automatically inferred from the data itself. Earlier approaches on self-supervised learning for images rely on manually designed pretext task, such as predicting relative position of image patches \cite{Doersch2015UnsupervisedVR}, inpainting missing parts of images \cite{pathak2016context}, and predicting angle of rotation applied to images \cite{gidaris2018unsupervised}. 

More recent approaches seek to directly impose consistency constraint on the representations. Contrastive learning \cite{chen2020simple, He2020MomentumCF} learns to minimize the distance between two similar representations (typically two augmented views of a same image) in a projected vector space, while maximizing the distance between two dissimilar representations. BYOL \cite{grill2020bootstrap} discards the notion of dissimilar representations by training a student network to predict the representations generated by an exponential moving average version of the student network, otherwise known as the teacher network. Barlow Twins \cite{Zbontar2021BarlowTS} applies the redundancy-reduction principle \cite{barlow1961possible} to self-supervised learning by forcing the cross-correlation matrix computed from two batches of representations to be close to the identity matrix. DINO \cite{Caron2021EmergingPI} performs self-distillation on two encoded views of the same image using a similar teacher-student setup from BYOL. These methods have largely closed the performance gap between self-supervised learning and supervised learning. However, our experiment results on retinal fundus datasets reveal that models pretrained on these consistency-based methods suffer performance drop when transferring to segmentation tasks with limited annotations. This might be partly attributed to the invariance property imposed by strong data augmentations, which can cause the learned representations to have poor visual grounding \cite{selvaraju2020casting}. Several works have been proposed to improve the localization ability of self-supervised learning, which include aligning feature maps of image crops \cite{roh2021spatially} and using unsupervised saliency maps \cite{selvaraju2020casting} or off-the-shelf region proposal tool \cite{bar2021detreg}. These unsupervised saliency methods are not directly applicable to the medical imaging setting due to the small subtleties and large variations of lesions typically encountered in medical images, which prompts us to explore the use of weak labels to inject information about lesions into image patches.

In the medical imaging domain, Model Genesis \cite{zhou2021models} outperforms ImageNet initialization in several 3D segmentation tasks with an image reconstruction-based objective designed specifically for 3D images. ConVIRT \cite{zhang2020contrastive} explores contrastive pretraining on image-text pairs and observe improvements in classification and zero-shot retrieval tasks. For MRI images, Chaitanya \textit{et al.} \cite{chaitanya2020contrastive} extends the contrastive learning objective to 3D images by partitioning each image volume into several partitions and forms positive pairs using partitions from the same location across different volumes. \textit{y}-Aware InfoNCE loss \cite{Dufumier2021ContrastiveLW} incorporates continuous image meta-data such as age information into contrastive pretraining, based on the intuition that images of similarly-aged patients should have similar representations. For retinal fundus images, Li \textit{et al.} endows learned representations with modality-invariant \cite{Li2020SelfSupervisedFL} and rotation-invariant \cite{Li2021RotationorientedCS} properties, through cross-modal image synthesis and rotation angle prediction task respectively.

% needed in second column of first page if using \IEEEpubid
%\IEEEpubidadjcol

\section{Methodology}
Briefly, our semi-weakly supervised pretraining framework consists of three main steps: 1) training a semi-supervised pseudo-labeler to propagate labels from a hand-labeled dataset to a large unlabeled dataset; 2) generating a semi-weakly annotated patch dataset for pretraining using CAMs from the pseudo-labeler; 3) pretraining on the generated dataset with multi-label contrastive learning objective. Below, we describe each step in more details.

\begin{figure*}[!t]
\centering
\includegraphics[width=6.5in]{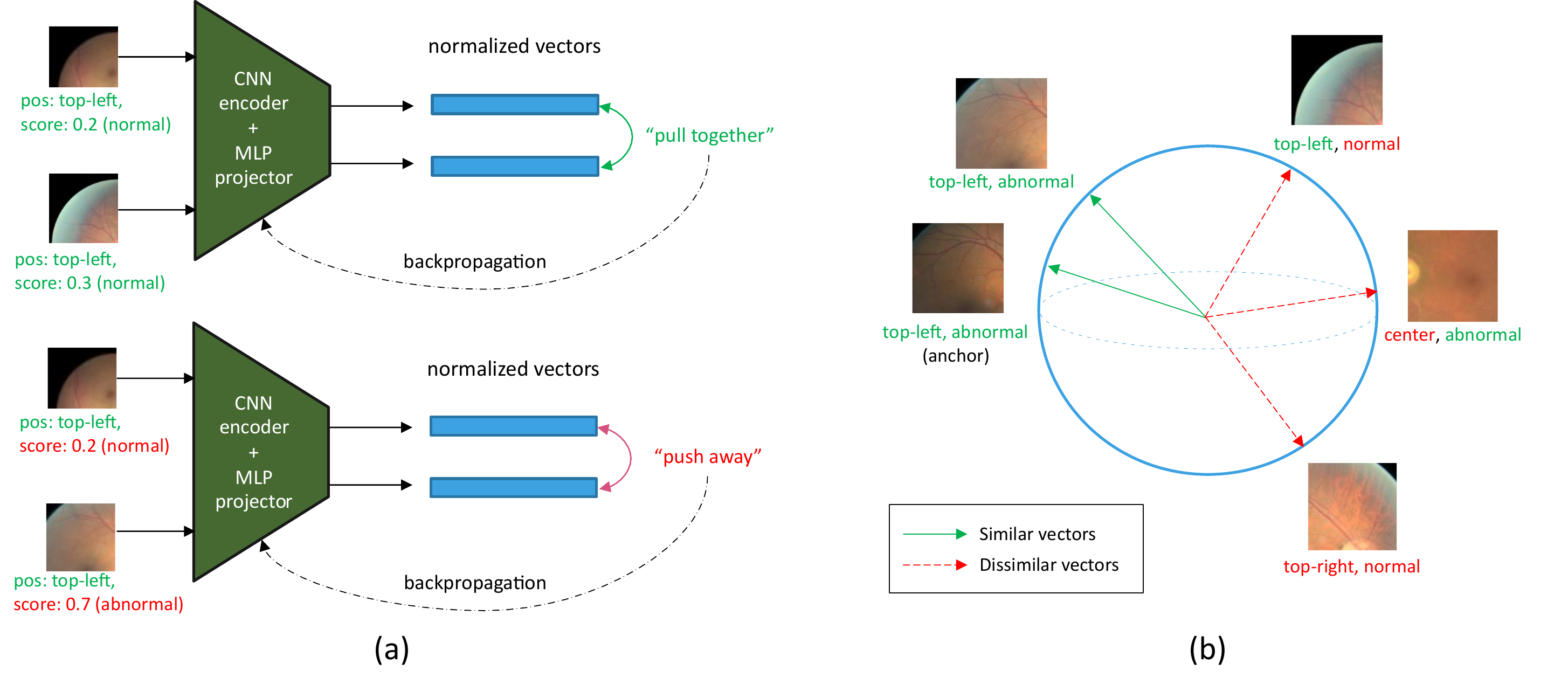}
\caption{Overview of the pretraining objective. (a) Contrastive learning on pairs of image patches. Positive pairs, defined as patches with the same position and abnormality labels (inferred from patch-wise lesion scores with threshold \(t\)) are encouraged to be close to each other in the projected vector space, while negative pairs with mismatched labels are pushed away from each other. Here, \(t\) is set to \(0.4\). (b) Illustration of the projected vector space after contrastive pretraining.}
\label{fig_swcl}
\end{figure*}

\subsection{Semi-supervised pseudo-labeler}
We train a semi-supervised binary CNN classifier using OIA-ODIR \cite{Li2020ABO} as the labeled dataset and images from Kaggle-EyePACS \cite{Cuadros2009EyePACSAA} as unlabeled dataset. Both OIA-ODIR and Kaggle-EyePACS consists of 10,000 and 88,702 fundus images from the left and right eyes of 5,000 and 44,351 patients respectively.\\ 

\textbf{Dataset preprocessing.} The left-right fundus pairs from OIA-ODIR were initially annotated with multi-hot labels from eight categories: normal, diabetic retinopathy, glaucoma, cataract, age-related macular degeneration, hypertension, myopia, and others. In addition, each individual fundus images were also annotated with diagnostic keywords in text form. To get an one-hot label for each fundus image, we convert the diagnostic keywords into a binary label by treating keywords associated with normal diagnosis as “normal” and other keywords as “abnormal”. Through this step, we convert the multi-label dataset into a binary classification dataset. Next, we discard the labels in Kaggle-EyePACS and treat the images as unlabeled because the original labels only describe the severity of DR and did not account for other disease types \cite{Wang2019RetinalAR}. Lastly, we resize all images from both labeled and unlabeled datasets to 448 pixels along the shorter side using bilinear sampling.\\

\textbf{Semi-supervised learning.} We adopt the self-supervised semi-supervised learning (S\textsuperscript{4}L) algorithm \cite{zhai2019s4l} as the learning algorithm. S\textsuperscript{4}L is a simple and effective algorithm with a self-supervised regularization term for unlabeled data. In particular, the learning objective of S\textsuperscript{4}L is given as:
\[ \min_\theta \mathcal{L}_{\ell}(D_\ell, \theta) + w\mathcal{L}_{u}(D_u, \theta), \]
where \(\mathcal{L}_\ell\) is a standard cross-entropy loss on the labeled dataset \(D_\ell\), \(\mathcal{L}_u\) is a loss defined for the unlabeled dataset \(D_u\), \(w\) is a non-negative scalar weight and \(\theta\) is the parameter of the model. We set \(w\) to \(1\) and implement \(\mathcal{L}_u\) as the hard triplet loss \cite{Hermans2017InDO} with a soft margin of \(0.5\). \\

\textbf{Training details.} Using the S4L algorithm, we train a ResNet-34 \cite{he2016deep} classifier (the ``pseudo-labeler'') from scratch on the preprocessed datasets. To obtain CAMs at a higher resolution (required for the next step), we reduce the stride in the last down-sampling block of ResNet-34 to 1 to prevent the feature maps from shrinking too much. At each optimization step, we generate four augmented views from each image and apply the loss function to each view. The data augmentation scheme includes random cropping to the size of 384\(\times\)384, random horizontal flipping and random color jittering \footnote{\url{https://pytorch.org/vision/stable/transforms.html#torchvision.transforms.ColorJitter}}. Following Zhai \textit{et al.} \cite{zhai2019s4l}, we also apply \(\mathcal{L}_u\) to the labeled images. We use the ``off-site'' split of OIA-ODIR as a hold-out validation set for hyperparameter tuning. The model is optimized using stochastic gradient descent (SGD) with Nesterov momentum, with a batch size of 256, a momentum of 0.9 and a weight decay of 0.001. The initial learning rate is set to 0.03 and decayed by a factor of 10 at epochs 140, 160, and 180. After training for 200 epochs, the classifier achieves an AUC-ROC score of 72\% on the hold-out validation set.

% An example of a double column floating figure using two subfigures.
% (The subfig.sty package must be loaded for this to work.)
% The subfigure \label commands are set within each subfloat command,
% and the \label for the overall figure must come after \caption.
% \hfil is used as a separator to get equal spacing.
% Watch out that the combined width of all the subfigures on a 
% line do not exceed the text width or a line break will occur.
%
%\begin{figure*}[!t]
%\centering
%\subfloat[Case I]{\includegraphics[width=2.5in]{box}%
%\label{fig_first_case}}
%\hfil
%\subfloat[Case II]{\includegraphics[width=2.5in]{box}%
%\label{fig_second_case}}
%\caption{Simulation results for the network.}
%\label{fig_sim}
%\end{figure*}
%
% Note that often IEEE papers with subfigures do not employ subfigure
% captions (using the optional argument to \subfloat[]), but instead will
% reference/describe all of them (a), (b), etc., within the main caption.
% Be aware that for subfig.sty to generate the (a), (b), etc., subfigure
% labels, the optional argument to \subfloat must be present. If a
% subcaption is not desired, just leave its contents blank,
% e.g., \subfloat[].

\subsection{Semi-weakly annotated patch dataset}
After the pseudo-labeler is trained, we pass the center crop of size 448\(\times\)448 of each image from the concatenation of OIA-ODIR and Kaggle-EyePACS to the pseudo-labeler and extract the class activation maps (CAMs) for both normal and abnormal class. CAMs can be interpreted as heatmaps that highlight the most discriminative regions for a particular class. Following Zhou \textit{et al.}, \cite{zhou2016learning}, we compute CAMs as the weighted sum of the feature maps of the last convolutional layer. Specifically, the value of a CAM at spatial location \((i, j)\) for a class \(c\), denoted as \(M_c(i, j)\), is computed by:
\[ M_c(i, j) = \sum_{k}{w^c_kf_k(i, j)}, \]
where \(k\) is the number of feature maps (channels) in the last convolutional layer, \(w^c_k\) is the weight of the fully connected layer connecting the \(k\)-th feature map to the output neuron of class \(c\), and \(f_k(i, j)\) is the value of the \(k\) feature map at spatial location \((i, j)\). For each image, we first compute CAMs for both the abnormal class, \( M_{a} \), and normal class, \( M_{n} \), before computing the normalized CAM for the abnormal class, \(\hat{M}_{a}\) via a softmax operation:
\[ \hat{M}_{a}(i, j) = \frac{\exp\left({M_{a}(i, j)}\right)}{\exp\left({M_{a}(i, j)}\right) + \exp\left({M_{n}(i, j)}\right)}. \]

To construct the patch dataset for pretraining, the images along with their normalized CAMs are cropped into five patches of equal size at five positions: top-left, top-right, center, bottom-left and bottom-right, as illustrated in Figure \ref{fig_gen}. To make the retinal structures aligned in each patch, we flip the right-eyed images and their CAMs horizontally before applying the crops. Each patch is then assigned two labels: a position label corresponding to the relative position of the crop and a lesion score label computed by averaging the values of cropped \(\hat{M}_{a}\). For the patches in OIA-ODIR whose image ground truth label is normal, we manually set the lesion scores to 0. A total of 493,470 semi-weakly annotated patches were generated through this annotation process.

\subsection{Semi-weakly supervised contrastive learning}
The goal of contrastive visual representation learning is to learn generalizable representations by discriminating positive image pairs against the negative pairs in a representation space. In self-supervised setting, positive pairs are typically defined as two differently augmented views of the same image and negative pairs are defined as any pairs of views that do not originate from the same image. Without the knowledge of image labels, this definition can result in false negative pairs where different images belonging to the same underlying class are treated as negative pairs. This causes their representations to be pushed away from each other in the representation space, leading to worse representations. To mitigate this issue, supervised contrastive learning \cite{NEURIPS2020_d89a66c7} adapts the self-supervised contrastive learning objective \cite{chen2020simple} to the fully supervised setting, by defining positive pairs as views from the same visual class. We further extend this supervised objective to the multi-label setting. Given \(N\) randomly sampled examples from a multi-label dataset with \(L\) labels, \(\{ x_k, y_{k,1},\dots, y_{k,L}\}_{k=1\dots N}\), where \(x_k\) is the \(k\)-th image and \(y_{k,\ell}\) is the \(\ell\)-th label of the \(k\)-th image, we first generate two augmented views for each image, such that each minibatch contains \(2N\) samples, \(\{ \tilde{x}_k, {\tilde{y}_{k,1},\dots, \tilde{y}_{k, L}}\}_{k=1\dots 2N}\), where \(\tilde{x}_{2k}\) and \(\tilde{x}_{2k-1}\) are two different views of \(x_k\) and \(\tilde{y}_{2k,\ell} = \tilde{y}_{2k-1,\ell} = y_{k,\ell}\) for \(\ell \in [1, L]\). To prevent clutter, we use \(\tilde{y}_k\) to denote the set of \(L\) labels for \(\tilde{x}_k\). Within each minibatch, we minimize the following loss function:

\small
\[\sum_{i=1}^{2N}{\frac{-1}{2N_{\tilde{y}_i} - 1}\sum_{j=1}^{2N}{\mathbbm{1}_{i \neq j} \cdot \mathbbm{1}_{\tilde{y}_i=\tilde{y}_j} \cdot \log\frac{\exp\left(z_i^\text{T} z_j / \tau\right)}{\sum_{k=1}^{2N}{\mathbbm{1}_{i \neq k} \cdot \exp\left(z_i^\text{T} z_k / \tau\right) } }  }} ,\]

where \( N_{\tilde{y}_i} \) is the number of instances in the minibatch that have the same labels as the anchor, \(i\), \(\mathbbm{1}_{condition}\) is an indicator function that returns \(1\) if \(condition\) evaluates to true and \(0\) if false, and \(\tau\) is a temperature parameter. We use the default value of \(0.1\) for \(\tau\) throughout all experiments. Vector \(z\) is the projection vector of \(\tilde{x}\), computed by a small projection network attached to the output of an encoder network (more details are provided in Section \ref{sec_arc}). The loss objective has the same form as the supervised contrastive learning objective, except that the class label \(y\) is a \(L\)-dimension vector instead of a scalar value and \(y_i = y_j\) if and only if each element in \(y_i\) matches \(y_j\), i.e., \(y_{i,\ell} = y_{j, \ell}\) for all \(\ell \in [1, L]\).

The multi-label formulation makes use of different types of labels presence in each image. As shown in Figure \ref{fig_swcl}, each patch of the retinal fundus image consists of two labels –- a position label and a lesion score label assigned during the semi-weakly annotation process. We convert the lesion scores into binary abnormality labels by applying a threshold, \(t\), on the scores, in which scores greater than or equal to \(t\) are assigned the abnormal label and scores lesser than \(t\) are assigned the normal label. We define positive pairs as patches with the same position and abnormality label. The intuition behind this definition is based on the observations that different position of the retina is usually associated with certain types of lesions, e.g., exudates surrounding the macula and hemorrhage near the blood vessels. Learning to encode positive pairs defined this way allows the encoder to capture features of position-specific lesions from the background structures, leading to a better representation quality. \\

\textbf{Practical consideration for patch alignment.} Although we have applied horizontal flipping to the right-eyed images to make sure they are roughly aligned with their left-counterparts, there is still a considerable misalignment among the images from Kaggle-EyePACS, which contains images from multiple screening sites captured under vastly different conditions, e.g., different lighting and zoom level. From our ablation study (see Section \ref{sec_abl}), we found that simply matching the position and abnormality labels across all patches would result in lower quality representations. Therefore, without resorting to complex image registration methods, we mitigate the misalignment issue by adding a patient-wise constraint to the definition of positive patches. Specifically, we treat two patches as positive pair only if they originated from the same patient while having matching position and abnormality labels. We implement this constraint by adding a patient identifier as the third label to each patch. Furthermore, to ensure there are sufficient inter-image positives in each minibatch, we sample the minibatch by pairing each left-eyed patch with the corresponding right-eyed patch of the same patient until there are \(N\) patches in the minibatch.

% An example of a floating table. Note that, for IEEE style tables, the
% \caption command should come BEFORE the table and, given that table
% captions serve much like titles, are usually capitalized except for words
% such as a, an, and, as, at, but, by, for, in, nor, of, on, or, the, to
% and up, which are usually not capitalized unless they are the first or
% last word of the caption. Table text will default to \footnotesize as
% the IEEE normally uses this smaller font for tables.
% The \label must come after \caption as always.
%
%\begin{table}[!t]
%% increase table row spacing, adjust to taste
%\renewcommand{\arraystretch}{1.3}
% if using array.sty, it might be a good idea to tweak the value of
% \extrarowheight as needed to properly center the text within the cells
%\caption{An Example of a Table}
%\label{table_example}
%\centering
%% Some packages, such as MDW tools, offer better commands for making tables
%% than the plain LaTeX2e tabular which is used here.
%\begin{tabular}{|c||c|}
%\hline
%One & Two\\
%\hline
%Three & Four\\
%\hline
%\end{tabular}
%\end{table}

\section{Experiments}

\begin{table*}[!t]
\caption{Details of the datasets used in fine-tuning experiments. \textsuperscript{\(\dagger\)}Following Sánchez \textit{et al.} \cite{Snchez2011EvaluationOA}, we treat Messidor as a coarse-grained classification dataset by grouping DR grade 0 and 1 as non-referable class and DR grade 2 and 3 as referable class. \textsuperscript{\(\ddagger\)}Following the conventional train-test split \cite{liu2019unsupervised}. *Accuracy when all predictions within a test sample are correct. (Abbreviations: DR - diabetic retinopathy; DME - diabetic macular edema; OD - optic disc; OC - optic cup)}
\label{tab_data}
\centering
\begin{tabular}{lllrrrl}
\toprule
 \multicolumn{1}{c}{Dataset} & \multicolumn{1}{c}{Task} & \multicolumn{1}{c}{\# classes} & \multicolumn{1}{c}{\# train} & \multicolumn{1}{c}{\# validation} & \multicolumn{1}{c}{\# test} & \multicolumn{1}{c}{Performance measure}  \\
\midrule
 Messidor \cite{decenciere_feedback_2014} &  Joint DR \& DME classification & DR: 2\textsuperscript{\(\dagger\)}, DME: 3 & 1196  & -   &  -                               & Joint accuracy\textsuperscript{*} \\
 IDRiD \cite{h25w98-18}                   &  Joint DR \& DME classification & DR: 5, DME: 3                              & 413   & -   &  103                             & Joint accuracy\textsuperscript{*} \\
 REFUGE-cls \cite{Orlando2020REFUGECA}    &  Galucoma classification        & 2                                          & 400   & 400 &  400                             & AUC-ROC                           \\
 DRIVE \cite{Staal2004RidgebasedVS}       &  Retinal vessel segmentation    & 2                                          & 20    & -   &  20\textsuperscript{\(\ddagger\)} & F1-score                          \\
 STARE \cite{Hoover1998LocatingBV}        &  Retinal vessel segmentation    & 2                                          & 10    & -   &  10\textsuperscript{\(\ddagger\)} & F1-score                          \\
 CHASE\_DB1 \cite{Fraz2012AnEC}           &  Retinal vessel segmentation    & 2                                          & 20    & -   &  8\textsuperscript{\(\ddagger\)} & F1-score                          \\
 REFUGE-seg \cite{Orlando2020REFUGECA}    &  Joint OD \& OC segmentation    & OD: 2, OC: 2                               & 400   & 400 & 400                              & Average F1-score                  \\
\bottomrule
\end{tabular}
\end{table*}

\begin{table*}[!t]
\caption{Transfer learning performance of different pretraining methods on the benchmark datasets. All methods are pretrained from scratch using the ResNet-50 (v2) backbone with group normalization and weight standardization. Each entry shows the mean evaluation score from five separate fine-tuning runs with different random seeds. The best overall scores are shown in \textbf{boldface} while the best scores obtainable by pretraining only on retinal fundus images are \underline{underlined}.}
\label{tab_results}
\centering
\begin{tabular}{lrrrrrrr}
\toprule
 \multicolumn{1}{c}{Pretrain method} & \begin{tabular}[c]{@{}c@{}}Messidor\\ (joint accuracy)\end{tabular} & \begin{tabular}[c]{@{}c@{}}IDRiD\\ (joint accuracy)\end{tabular} & \begin{tabular}[c]{@{}c@{}}REFUGE-cls\\ (AUC-ROC)\end{tabular} & \begin{tabular}[c]{@{}c@{}}DRIVE\\ (F1-score)\end{tabular} & \begin{tabular}[c]{@{}c@{}}CHASE\_DB1\\ (F1-score)\end{tabular} & \begin{tabular}[c]{@{}c@{}}STARE\\ (F1-score)\end{tabular} & \begin{tabular}[c]{@{}c@{}}REFUGE-seg\\ (average F1.score)\end{tabular} \\
\midrule
random                                  & 60.96	             & 32.62	          & 85.27	                   & 76.83	           & 76.52	                     & 74.34	           & 85.66             \\
\midrule                                                                                                                                                                                                 
SimCLR \cite{chen2020simple}            & 66.81	             & 45.83	          & 93.19	                   & 77.11	           & 77.64	                     & 75.02	           & 88.01             \\
BYOL \cite{grill2020bootstrap}          & 67.14	             & 38.25	          & 91.96                      & 76.15	           & 76.32	                     & 72.52	           & 85.66             \\
DINO \cite{Caron2021EmergingPI}         & 64.46	             & 40.19	          & 92.59	                   & 76.30	           & 77.15	                     & 74.54	           & 84.74             \\
\midrule                                                                                                                                                                                                 
OIA-ODIR                                & 64.05	             & 38.64	          & 86.68	                   & 76.29	           & 77.46                       & 76.96	           & 83.95             \\
Kaggle-EyePACS                          & 67.06	             & 46.41	          & 91.80	                   & 77.99	           & 77.92	                     & 77.41	           & 88.30             \\
BiT-S \cite{Kolesnikov2020BigT}         & 76.59	             & \textbf{54.76}	  & 91.99	                   & \textbf{78.88}    & 78.19                       & 78.88	           & 90.70             \\
BiT-M \cite{Kolesnikov2020BigT}         & \textbf{79.35}     & 53.59	          & 93.73	                   & 77.93	           & 78.50      	             & \textbf{79.37}	   & \textbf{90.91}    \\
\midrule                                                                                                                                                                             
CE-multitask                            & 66.56	             & 45.05	          & 90.97	                   & \underline{78.24} & \underline{\textbf{78.57}}	 & 77.24     	       & 89.30             \\
SWCL-image                              & 70.65              & 48.54              & 92.93                      & 78.05             & 78.34                       & \underline{78.44}   & \underline{90.01} \\
SWCL                                    & \underline{71.57}	 & \underline{51.26}  & \underline{\textbf{95.86}} & 78.08	           & \underline{\textbf{78.61}}	 & \underline{78.45}   & 89.65             \\ 
\bottomrule
\end{tabular}
\end{table*}

\subsection{Datasets}
To evaluate the quality of learned representations, we fine-tune the models in an end-to-end fashion on seven retinal fundus datasets with very limited annotations. The details of each dataset are summarized in Table \ref{tab_data}. We resize each image from all classification datasets until the shorter side is 350 pixels. For segmentation datasets, we only resize the images from REFUGE-seg to 514\(\times\)514 pixels.

\subsection{Experiment setups}
\textbf{Baselines}. We compare our method (SWCL) against other weight initialization methods, including:
\begin{itemize}
    \item random initialization (no pretraining) – \textbf{random}
    \item self-supervised baselines (pretrained on the concatenation of OIA-ODIR and Kaggle-EyePACS without labels) \\– \textbf{SimCLR \cite{chen2020simple}, BYOL \cite{grill2020bootstrap}, DINO \cite{Caron2021EmergingPI}}
    \item supervised baselines (pretrained on hand-labeled datasets) \\– \textbf{OIA-ODIR, Kaggle-EyePACS, BiT-S \cite{Kolesnikov2020BigT}, Bit-M \cite{Kolesnikov2020BigT}}
    \item multitask cross-entropy baseline (pretrained on retinal-SWAP) – \textbf{CE-multitask}
    \item image-level SWCL baseline (pretrained on retinal-SWAP, but the patch-level abnormality labels are replaced with image-level labels) – \textbf{SWCL-image}
\end{itemize}

We obtain the pretrained weights for BiT-S and BiT-M from the official BiT repository\footnote{\url{https://github.com/google-research/big_transfer}}. Both models were pretrained on large-scale image classification tasks, in which BiT-S was trained on the ILSVRC-2012 \cite{Russakovsky2015ImageNetLS} dataset consisting of 1.3M images while BiT-M was trained on the ImageNet-21k \cite{Deng2009ImageNetAL} dataset consisting of 14M images. We use both of these pretrained models as the upper bound baselines for large-scale supervised pretraining. \\

\label{sec_arc}
\textbf{Model architecture.} For the main experiments, we use the BiT \cite{Kolesnikov2020BigT} variant of  ResNet-50x1 as the encoder backbone. This architecture is an improved version of the ResNet-v2 architecture \cite{he2016identity} with all its Batch Normalization \cite{ioffe2015batch} layers replaced with Group Normalization \cite{Wu2018GroupN} and Weight Standardization \cite{Qiao2019WeightS}. For the projection network, we use a 2-layer multi-layer perceptron (MLP) with an output dimension of 128 (same as the one used in SimCLR \cite{chen2020simple}). When fine-tuning on downstream tasks, the projector network is discarded and replaced with either task-specific linear classification layers (for classification tasks) or a segmentation network (for segmentation tasks). We use the DeepLabv3+ \cite{chen2018encoder} architecture as the segmentation network. \\

\textbf{Data augmentation scheme.} The pretraining stage follows the same data augmentation scheme used in SimCLR \cite{chen2020simple} and BYOL \cite{grill2020bootstrap}, which includes random cropping, random horizontal flip, random color jittering, random gray scale, Gaussian blurring and solarization. In the fine-tuning experiments, we use random crop (224\(\times\)224 for classification tasks, 384\(\times\)384 for segmentation tasks), random horizontal flip and random gray scale in the training stage. During inference, we resize the images to 256 pixels along the shorter side before taking the 224\(\times\)224 center crops as inputs for the classification tasks; for segmentation tasks, we stich together patches of 384\(\times\)384 segmentation masks obtained in a sliding window fashion. After applying data augmentations, the color channels of all images are normalized by the mean and standard deviation of colors computed on the Kaggle-EyePACS dataset. \\

\label{sec_hyp}
\textbf{Pretraining details.} All models are pretrained from randomly initialized encoder-projector networks using the AdamW \cite{Loshchilov2019DecoupledWD} optimizer. For our main method (SWCL), we set the base learning rate to 0.001 and train the model for 40 epochs with a batch size of 496 and a weight decay of 1e-4. The learning rate is linearly increased to the base value in the first 5 epochs before gradually decaying towards 0 using a cosine scheduler. For the baseline methods, we separately tune the values of base learning rate and weight decay. Since retinal-SWAP contains five times more training examples than the image-level datasets, we train the self-supervised baselines for 200 epochs such that the number of optimization steps is similar. \\

\textbf{Fine-tuning details.} For datasets without validation split (i.e., IDRiD, DRIVE, STARE, CHASE\_DB1), we randomly select 20\% of the samples from the training split as validation split; for dataset without both validation and test split (i.e., Messidor), we report results from 5-fold cross validation. During fine-tuning, we initialize each neural network with the parameters of a pretrained network and optimize the loss using SGD with Nesterov momentum with a momentum of 0.9. Models in classification tasks are trained with a batch size of 64 for up to 120 epochs while models in segmentation tasks are trained with a batch size of 8 for up to 300 epochs. For each task and initialization method, we perform grid search to select the best learning rate and weight decay parameters. After selecting the best-performing hyperparameters, we train a new model for each task using the selected parameters on the train split with early stopping applied to the validation split. We repeat each experiment five times using five different random seeds and report the performance measure on the test split.

\begin{figure*}[!t]
\centering
\includegraphics[width=6.5in]{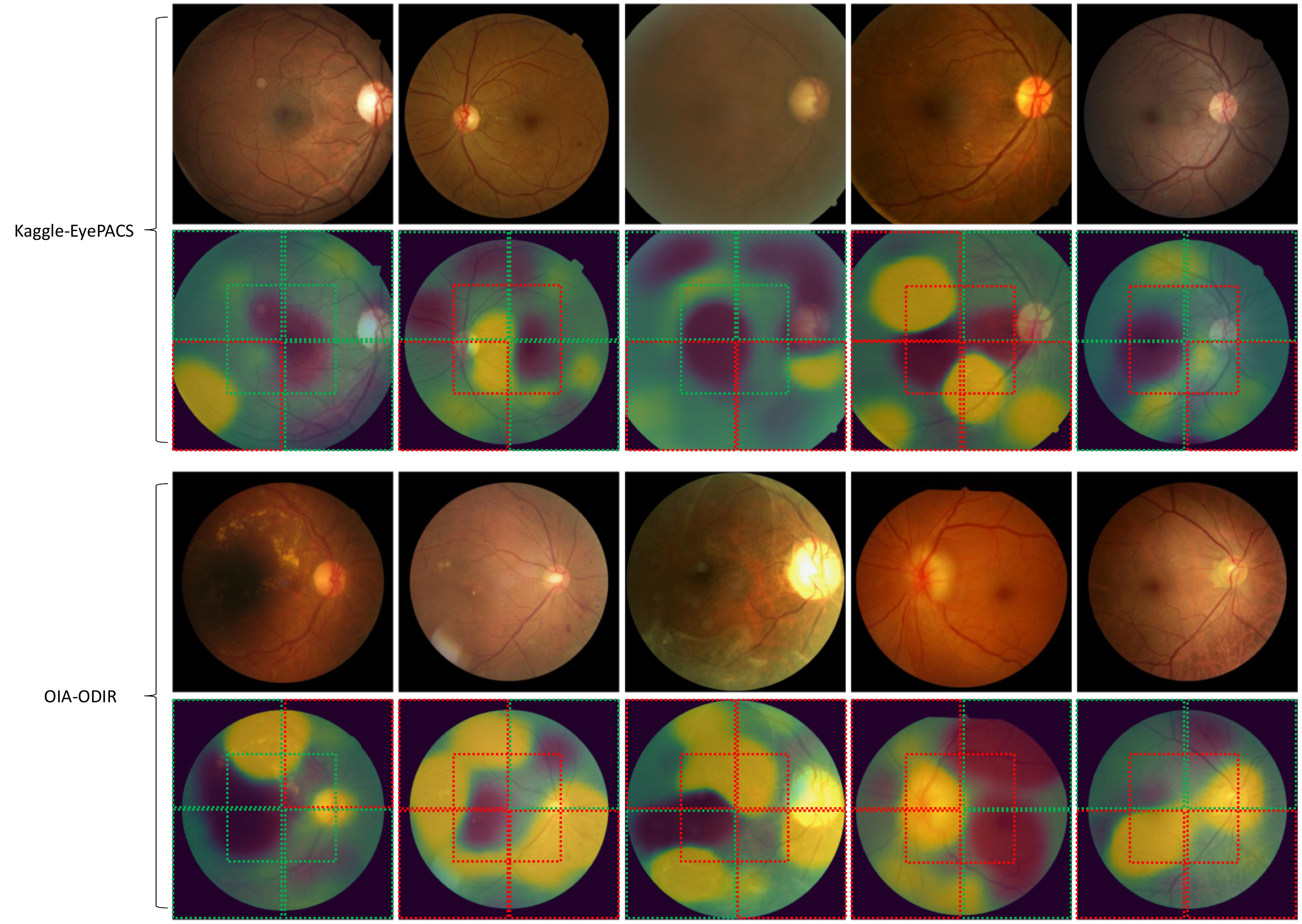}
\caption{Visualizations of the CAMs of unlabeled images computed for the abnormal class from the pseudo-labeler. The first and third rows show the unlabeled images from Kaggle-EyePACS and OIA-ODIR respectively; the second and fourth rows show the CAMs (upscaled to the size of the image) overlaid on top of the images. The dotted bounding boxes correspond to the five patches taken from each image; boxes with green outline are normal patches with \(t < 0.4\) and boxes with red outline are abnormal patches with \(t >= 0.4\). Best viewed in color.}
\label{fig_cam}
\end{figure*}

\subsection{Main results}
\label{sec_results}
The transfer learning performance measured on the test sets are reported in Table \ref{tab_results}. Below we highlight several key observations from the table. \\

\textbf{All pretraining methods outperforms training from scratch in classification tasks.} Under very low data regime, models initialized from pretrained weights consistently outperform a randomly initialized model on all three disease classification tasks, with improvements as large as 22.14\% joint accuracy on the IDRiD benchmark with BiT-S initialization. This suggests that pretraining on either labeled or unlabeled data in general can provide a great starting initialization for downstream classification tasks. The number of in-domain training examples matters as well — the Kaggle-EyePACS baseline which was trained on a larger set of labeled retinal fundus images achieve comparably better downstream performance than the OIA-ODIR baseline. On the contrary, this is not the case in ImageNet pretraining where BiT-S which is pretrained on the smaller subset of ImageNet outperforms BiT-M on two benchmarks (i.e., IDRiD, DRIVE). \\

\textbf{SWCL improves segmentation performance.} Despite obtaining good transfer performance in classification tasks, models pretrained using self-supervised methods have poor transfer performance when it comes to dense prediction tasks. For example, BYOL and DINO achieve slightly worse or comparable segmentation performance to the random baseline on the DRIVE, STARE and REFUGE-seg benchmarks. The drops in segmentation performance can be partly attributed to the invariance properties imposed at the image-level \cite{roh2021spatially}. During pretraining, self-supervised objectives force different augmented patches of the same image to have consistent representations regardless of the relative position of the patches. This reduces the amount of position-related information captured by the encoder, which in turn causes performance drop in tasks that depend on such information, i.e., dense prediction tasks. In contrast, with the guidance of position-specific lesion information and patch alignment, SWCL is able to learn representations that benefit both classification and segmentation tasks under the same data augmentation scheme. \\

\textbf{SWCL outperforms in-domain cross-entropy objectives.} Compared to pretraining with standard cross-entropy objectives on single-task hand-labeled datasets (OIA-ODIR, Kaggle-EyePACS) and the generated multi-task retinal-SWAP dataset (CE-multitask), SWCL consistently achieves better performance across almost all benchmark datasets. Notably, SWCL improves upon the next best method (Kaggle-EyePACs) by 4.51\% and 4.85\% joint accuracy on the Messidor and IDRiD benchmark, respectively, and 4.06\% AUC-ROC on the REFUGE-cls benchmark. Interestingly, CE-multitask only surpass Kaggle-EyePACS on three benchmarks (i.e., DRIVE, CHASE\_DB1, REFUGE-seg) despite having been trained on patches from Kaggle-EyePACS and OIA-ODIR. This highlights the shortcomings of cross-entropy objective in representation learning and illustrates the effectiveness of our proposed multi-label contrastive learning objective. \\

\textbf{Replacing semi-weakly annotated patch labels with image-level labels.} We also experiment with another way of assigning patch-wise abnormality labels by replacing the semi-weakly annotated labels with image-level labels from the original datasets. Similar to SWCL, models initialized with weights trained on the replaced abnormality labels (SWCL-image) surpass both self-supervised and in-domain cross-entropy pretraining, which shows that the addition of position information and contrastive learning plays a crucial role in representation learning for retinal fundus images. By injecting patch-specific lesion information extracted from the CAMs of a semi-supervised pseudo-labeler, our method (SWCL) achieves even better performance on the classification benchmarks while maintaining competitive segmentation performance. \\

\textbf{SWCL closes the gap with large-scale pretraining.} With improved architectural design (i.e., group normalization \cite{Wu2018GroupN}, weight standardization \cite{Qiao2019WeightS}) and large hand-labeled datasets, BiT-S and BiT-M achieve remarkable transfer learning performance on a wide variety of natural image classification tasks \cite{Kolesnikov2020BigT}. Our experiment results show that the dominant transfer learning performance of BiT-S and BiT-M is also applicable to retinal fundus images. In five out of seven benchmarks (i.e., Messidor, IDRiD, DRIVE, STARE, REFUGE-seg), models initialized with weights from BiT-S or BiT-M achieve the best overall performance. By training on a relatively small semi-weakly annotated dataset, SWCL closes the performance gap in these five benchmarks and outperforms the large-scale pretraining methods in the remaining two benchmarks (i.e., REFUGE-cls, CHASE\_DB1). \\

% Note that the IEEE does not put floats in the very first column
% - or typically anywhere on the first page for that matter. Also,
% in-text middle ("here") positioning is typically not used, but it
% is allowed and encouraged for Computer Society conferences (but
% not Computer Society journals). Most IEEE journals/conferences use
% top floats exclusively. 
% Note that, LaTeX2e, unlike IEEE journals/conferences, places
% footnotes above bottom floats. This can be corrected via the
% \fnbelowfloat command of the stfloats package.

\subsection{Ablation study}
\label{sec_abl}
We conduct ablation studies for different values of \(t\) and different labelling schemes in the multi-label contrastive objective. For each configuration, we pretrain a ResNet-18 model from scratch and report the fine-tuning performance on the validation split of REFUGE-cls. Each model is trained for 40 epochs with a batch size of 512. The rest of the hyperparameters follow the values described in Section \ref{sec_hyp}. The ablation results are given in Table \ref{tab_abl}. \\

\begin{table}[ht]
\caption{Ablation on different threshold values and labelling schemes. Performance measure is given in AUC-ROC score evaluated on the validation split of REFUGE-cls. We also include a random initialization baseline (Random init.) and a supervised baseline pretrained on ILSVRC-2012 (ILSVRC-2012 init.).}
\label{tab_abl}
\centering
\begin{tabular}{lr}
\toprule
\multicolumn{1}{c}{Configuration} & \multicolumn{1}{c}{AUC-ROC} \\
\midrule
Random init.	                        & 71.88          \\
ILSVRC-2012 init.	                    & 88.18          \\
\midrule
\(t = 0.3\)	                            & 95.03          \\
\(t = 0.4\) (default)	                & \textbf{96.52} \\
\(t = 0.5 \)	                        & 92.91          \\
\midrule
Remove position label	                & 95.78          \\
Remove abnormality label	            & 93.00          \\
Remove patient label	                & 95.68          \\
Remove position and abnormality label	& 94.43          \\
Remove position and patient label	    & 87.51          \\
Remove abnormality and patient label	& 85.85          \\
\bottomrule
\end{tabular}
\end{table}

\textbf{Optimal threshold.} Under the full multi-label scheme for contrastive representation learning (i.e., with position, abnormality, and patient labels), the AUC-ROC score improves by 1.49\% and peaks at 96.52\% as \(t\) is increased from 0.3 to 0.4. When \(t\) is increased to 0.5, the classification performance drops to 92.91\% (a decrease of 3.61\%), due to the increase in false negatives as more patches are labeled as normal. \\

\textbf{Label matching scheme.} We gradually remove the labels from the multi-label contrastive learning objective to assess how each label impact the representation quality. When only position or patient label is removed, the performance drop is lesser than 1\%. In contrast, when the abnormality label is removed, the performance drop increases to 3.52\%, demonstrating that generic lesion information plays an important role in learning representations that can generalize to other diseases. However, if we remove all labels except the abnormality label, the AUC-ROC score drops below the performance of a ILSVRC-2012 baseline (88.18\%) to 87.51\%. Coupled with the performance of in-domain pretraining baselines discussed in Section \ref{sec_results}, the results suggests that abnormality information alone is not sufficient for generic visual representation learning, indicating the importance of position alignment.

\subsection{Analysis}
\textbf{Class activation maps.} We visualize the CAMs extracted for the abnormal class from the pseudo-labeler for randomly selected images in Figure \ref{fig_cam}. From the figure, we can see that the pseudo-labeler assigns high activation values to regions with noticeable exudates (e.g., towards the center of the second image in the second row), as well as the swollen optic disc appearing in the center of the fourth image in the last row. Compared to image-level labels, the patch-level lesion scores derived from CAM patches prevent normal patches from being classified as abnormal when the image-level label is abnormal. These position-specific labels can provide a more informative training signals for the contrastive learning objective as the model has to discriminate between normal and abnormal patches from whole images. \\

\begin{figure}[ht]
\centering
\includegraphics[width=3.3in]{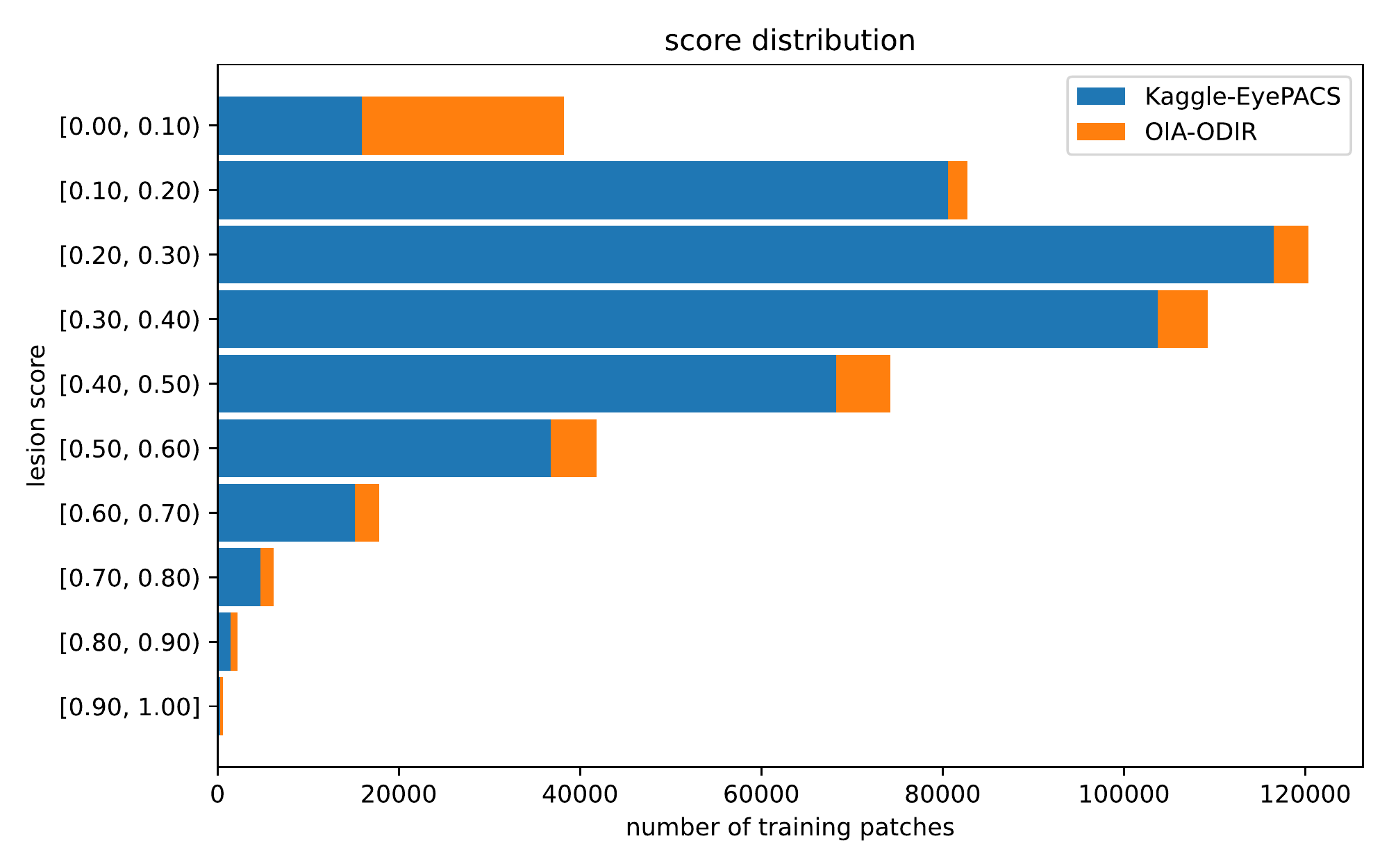}
\caption{Distribution of lesion scores in the semi-weakly annotated patch dataset.}
\label{fig_dist}
\end{figure}

\textbf{Lesion scores distributions.} Figure \ref{fig_dist} plots the distribution of the lesion scores computed by the pseudo-labeler on the concatenation of OIA-ODIR and Kaggle-EyePACS. The lesion scores among the unlabeled images form a bell-shaped curve with peaks around the interval between 0.2 and 0.3. The optimal score threshold, i.e., \(t = 0.4\), produces 142,957 abnormal patches, which accounts for around 29\% (out of 493,470 patches) of the patches used in the pretraining stage. Majority of the patches from OIA-ODIR has scores between 0 to 0.1 because we have assumed that normal patches (whose image-level label is normal) contain no lesion region and explicitly set the score of each normal patch to 0.

\section{Conclusion}
We propose a semi-weakly supervised representation learning framework for retinal fundus images. Through semi-supervised training, we propagate existing image-level annotations from a small hand-labeled data to another large unlabeled dataset in the form of class activation maps, which are then used to generate a large patch-level pretraining dataset with semi-weak annotations. Together with multi-label contrastive learning objective, our framework is competitive with large-scale supervised ImageNet pretraining and can produce highly transferable representations for a wide variety of downstream tasks such as joint disease classification and anatomical structure segmentation. Our pretraining framework can also be easily extended to other imaging modalities with high structural similarities among images, e.g., CT scans and radiographic images.

% if have a single appendix:
%\appendix[Proof of the Zonklar Equations]
% or
%\appendix  % for no appendix heading
% do not use \section anymore after \appendix, only \section*
% is possibly needed

% use appendices with more than one appendix
% then use \section to start each appendix
% you must declare a \section before using any
% \subsection or using \label (\appendices by itself
% starts a section numbered zero.)
%

% \appendices
% \section{Proof of the First Zonklar Equation}
% Appendix one text goes here.

% you can choose not to have a title for an appendix
% if you want by leaving the argument blank
% \section{}
% Appendix two text goes here.

% use section* for acknowledgment
% \section*{Acknowledgment}

% The authors would like to thank...

% Can use something like this to put references on a page
% by themselves when using endfloat and the captionsoff option.
\ifCLASSOPTIONcaptionsoff
  \newpage
\fi

% trigger a \newpage just before the given reference
% number - used to balance the columns on the last page
% adjust value as needed - may need to be readjusted if
% the document is modified later
%\IEEEtriggeratref{8}
% The "triggered" command can be changed if desired:
%\IEEEtriggercmd{\enlargethispage{-5in}}

% references section

% can use a bibliography generated by BibTeX as a .bbl file
% BibTeX documentation can be easily obtained at:
% http://mirror.ctan.org/biblio/bibtex/contrib/doc/
% The IEEEtran BibTeX style support page is at:
% http://www.michaelshell.org/tex/ieeetran/bibtex/
\bibliographystyle{IEEEtran}
% argument is your BibTeX string definitions and bibliography database(s)
%\bibliography{IEEEabrv,../bib/paper}
%
% <OR> manually copy in the resultant .bbl file
% set second argument of \begin to the number of references
% (used to reserve space for the reference number labels box)
% \begin{thebibliography}{1}

% \bibitem{IEEEhowto:kopka}
% H.~Kopka and P.~W. Daly, \emph{A Guide to \LaTeX}, 3rd~ed.\hskip 1em plus
%   0.5em minus 0.4em\relax Harlow, England: Addison-Wesley, 1999.

% \end{thebibliography}
\bibliography{bibtex.bib}

% biography section
% 
% If you have an EPS/PDF photo (graphicx package needed) extra braces are
% needed around the contents of the optional argument to biography to prevent
% the LaTeX parser from getting confused when it sees the complicated
% \includegraphics command within an optional argument. (You could create
% your own custom macro containing the \includegraphics command to make things
% simpler here.)
%\begin{IEEEbiography}[{\includegraphics[width=1in,height=1.25in,clip,keepaspectratio]{mshell}}]{Michael Shell}
% or if you just want to reserve a space for a photo:

% \begin{IEEEbiography}{Michael Shell}
% Biography text here.
% \end{IEEEbiography}

% if you will not have a photo at all:
% \begin{IEEEbiographynophoto}{John Doe}
% Biography text here.
% \end{IEEEbiographynophoto}

% insert where needed to balance the two columns on the last page with
% biographies
%\newpage

% \begin{IEEEbiographynophoto}{Jane Doe}
% Biography text here.
% \end{IEEEbiographynophoto}

% You can push biographies down or up by placing
% a \vfill before or after them. The appropriate
% use of \vfill depends on what kind of text is
% on the last page and whether or not the columns
% are being equalized.

%\vfill

% Can be used to pull up biographies so that the bottom of the last one
% is flush with the other column.
%\enlargethispage{-5in}

% that's all folks
\end{document}